\documentclass[10pt,twocolumn,letterpaper]{article}

\usepackage{cvpr} % [review]
\usepackage{times}
\usepackage{epsfig}
\usepackage{graphicx}
\usepackage{amsmath}
\usepackage{amssymb}

\begin{document}

%%%%%%%%% TITLE
\title{\vspace{-35pt}Deep Visual-Genetic Biometrics for Taxonomic 
 Classification of Rare Species\vspace{-10pt}}

\author{Tayfun Karaderi\\
\small Dept of Computer Science\\
\small University of Bristol\\
{\tt\footnotesize vm19402@bristol.ac.uk}
\and
Tilo Burghardt\\
\small Dept of Computer Science\\
\small University of Bristol\\
{\tt\footnotesize tilo@cs.bris.ac.uk}
\and
Rapha\"{e}l Morard\\
\small MARUM\\
\small University of Bremen\\
{\tt\footnotesize rmorard@marum.de}
\and
Daniela N. Schmidt\\
\small School of Earth Sciences\\
\small University of Bristol\\
{\tt\footnotesize d.schmidt@bristol.ac.uk}\vspace{-10pt}
}

\maketitle
\thispagestyle{empty}

%%%%%%%%% ABSTRACT
\begin{abstract} \vspace{-5pt}
  Visual as well as genetic biometrics are routinely employed to identify species and individuals in biological applications. However, no attempts have been made in this domain to computationally enhance visual classification of rare classes with little image data via genetics. In this paper, we thus propose aligned visual-genetic learning as a new application domain with the aim to implicitly encode cross-modality associations for improved performance.  We demonstrate for the first time that such alignment can be achieved via deep embedding models and that the approach is directly applicable to boosting long-tailed recognition (LTR), particularly for rare species. We experimentally demonstrate the efficacy of the concept via application to microscopic imagery of 30k+ planktic foraminifer shells across 32 species when used together with independent genetic data samples. Most importantly for practitioners, we show that visual-genetic alignment can significantly benefit visual-only recognition of the rarest species. Technically, we pre-train a visual ResNet50 deep learning model using triplet loss formulations to create an initial embedding space. We re-structure this space based on genetic anchors embedded via a Sequence Graph Transform (SGT) and linked to visual data by cross-domain cosine alignment. We show that an LTR approach improves the state-of-the-art across all benchmarks and that adding our visual-genetic alignment improves per-class and particularly rare tail class benchmarks significantly further. Overall, visual-genetic LTR training raises rare per-class accuracy from 37.4\% to benchmark-beating 59.7\%. We conclude that visual-genetic alignment can be a highly effective tool for complementing visual biological data containing rare classes. The concept proposed may serve as an important future tool for integrating genetics and imageomics towards a more complete scientific representation of taxonomic spaces and life itself. Code, weights, and data splits are published for full reproducibility.
\end{abstract}

\begin{figure}
  \begin{center}\vspace{5pt}
    \includegraphics[width=85mm, height=80mm]{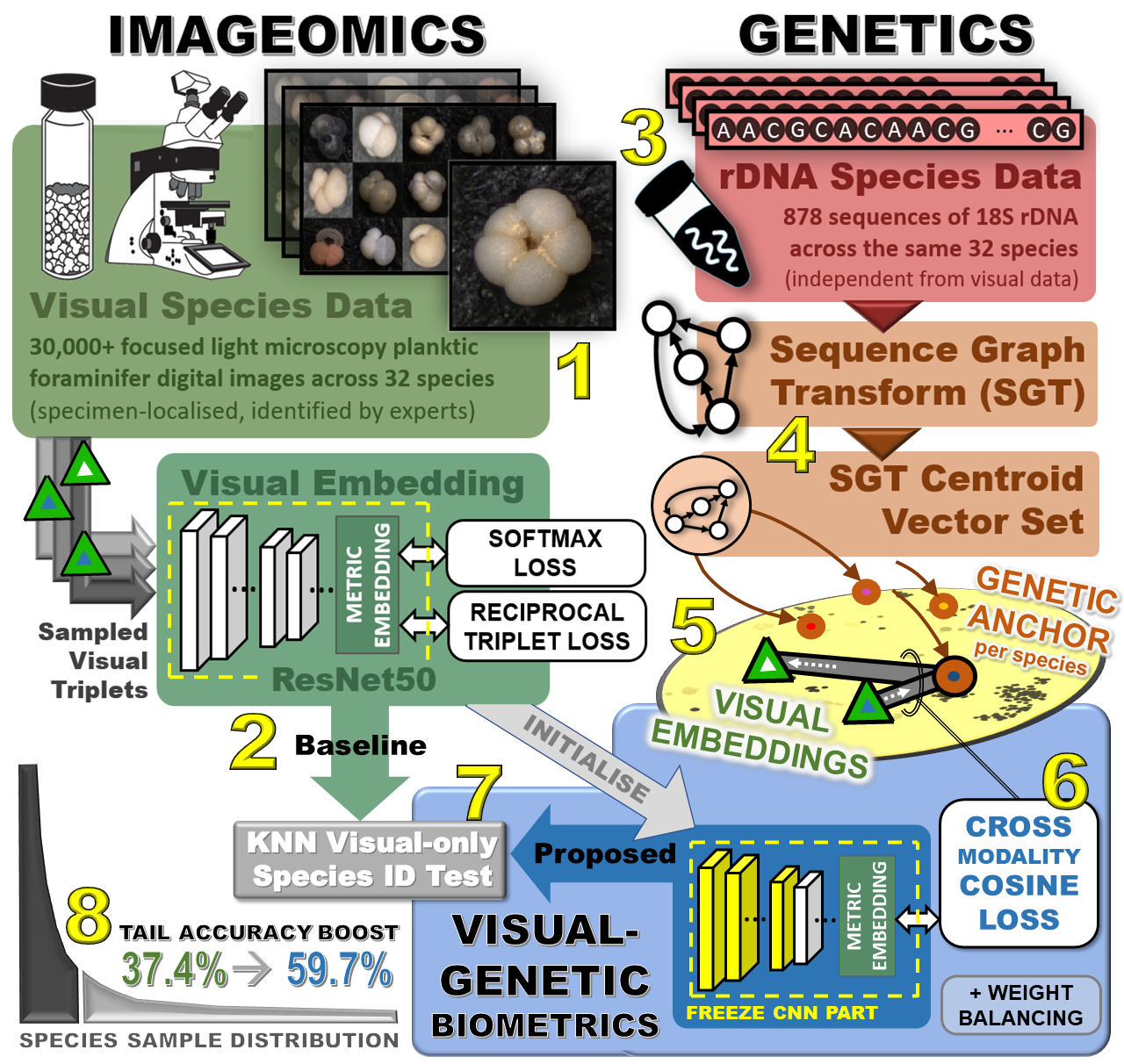}\vspace{-15pt}
  \end{center}
  \caption{\textbf{Visual-Genetic Co-Learning Architecture.} This work combines imageomics with genetics to improve on visual biometric recognition of particularly rare species in biology. \textbf{\textit{(1)}} 30k+ images of 32 species from the EndlessForams dataset are used to train \textbf{\textit{(2)}} a traditional metric deep learning baseline for species classification. However, we also use \textbf{\textit{(3)}} independent rDNA sequence data \textbf{\textit{(4)}} transformed into the same embedding space via SGT. Each species can now also be represented via genetic anchor information. \textbf{\textit{(5)}} Cross-modality triplets of a genetic anchor and a positive and negative visual embedding are co-used via \textbf{\textit{(6)}} the cosine transform to learn a visual-genetic species space adjusting late layers of the visual embedding. We show that \textbf{\textit{(7)}} cosine KNN visual-only testing of such a network when weight-balanced can \textbf{\textit{(8)}} significantly improve performance, particularly for rare species.\vspace{-15pt}}
  \label{title-figure}
\end{figure}
\vspace{-15pt}

%%%%%%%%% BODY TEXT
\section{Introduction}\vspace{-4pt}
\subsection{Motivation}\vspace{-4pt}
\textbf{Visual, Genetic, and Long-Tailed Data in Biology.} Both genetic and visual biometrics are extensively utilised to support species and individual identification in biological applications~\cite{kuehl2013,tuia2022,valentini2009}. Yet, modalities are usually learned and processed independently without explicitly considering cross-modal information. In how far information from genetics of a species can assist classification of visuals of the phenotype is of particular practical interest given that visual source or training data for imageomics~\cite{imageomics,stennett2022towards,tuia2022,yang2023dynamic} may be prohibitively limited for some classes (e.g. visual samples for very rare species). In fact, the distribution of most biological species datasets~\cite{EndlessForams,perrett2023,inaturalist} follows a ‘long-tailed’ pattern or at least contain many rare classes. Thus, models trained on such data often struggle accurately to encode and consequently recognise less common species. 

\textbf{Cross-Modal Taxonomic Information.} In living organisms, the relationship between taxa is traditionally determined by their genetics. However, sister taxa that are closely related commonly share morphological features observable via imaging techniques too. Consequently, visual and genetic feature distances between species as well as their orientation in any overarching, cross-domain feature space should be related to some degree. We therefore hypothesise that enriching imageomic representation spaces via information transfer from genetics may enhance deep visual species representation models particularly when the latter is built under long-tailed training data limitations. 

\textbf{Deep Visual-Genetic Embedding.} In this paper and following the above line of argument, we explore enriching deep imageomics for taxonomic species classification with independently sourced genetic information in order to improve visual-only species recognition performance for long-tailed datasets. Fig.~\ref{title-figure} provides a schematic overview of the proposed approach and how it combines imageomics with genetics to improve visual classification of rare species. In particular, we propose utilising relative orientation information from rDNA (ribosomal genes DNA) embeddings to optimise visual embeddings. Technically, state-of-the-art (SOTA) triplet loss formulations~\cite{Tayfun} for learning metric visual classification spaces are expanded across modalities in a second learning stage that uses rDNA anchors and cosine similarity metrics to draw in additional information from the genetic domain. We test the approach on the challenging task of identifying planktic foraminifer species at scale, which is of critical importance for paleoclimatology.

\subsection{Paper Contributions}\vspace{-3pt}
To the best of our knowledge, this work employs rDNA information to guide the orientations of deep visual embeddings for species recognition for the first time. Our key contributions are as follows:\vspace{-6pt}
\begin{itemize}
    \item  \textbf{Concept.} We propose a new type of modality integration for imageomics combining visual and genetic information in one metric space usable for inference.\vspace{-6pt}
    \item  \textbf{Implementation.} We provide a deep transfer learning framework that implements the new concept for biological applications and publish the full source code of this visual-genetic co-learning architecture.\vspace{-6pt}
    \item  \textbf{Experimentation.} We demonstrate that the proposed implementation achieves state-of-the-art accuracy results whilst significantly boosting tail class recognition performance for a visual species recognition task on a large 30k+ example image set covering 32 species of difficult to identify planktic foraminifers. We provide all weights and data splits for full reproducibility.\vspace{-6pt}
    \item \textbf{Analysis.} We structurally analyse the novel visual-genetic spaces created and interpret as well as visualise effects of cross-modal alignment.
\end{itemize}

\section{Related Work}\vspace{-3pt}

\begin{figure}
  \begin{center}\vspace{-8pt}
    \includegraphics[width=80mm,height=55mm]{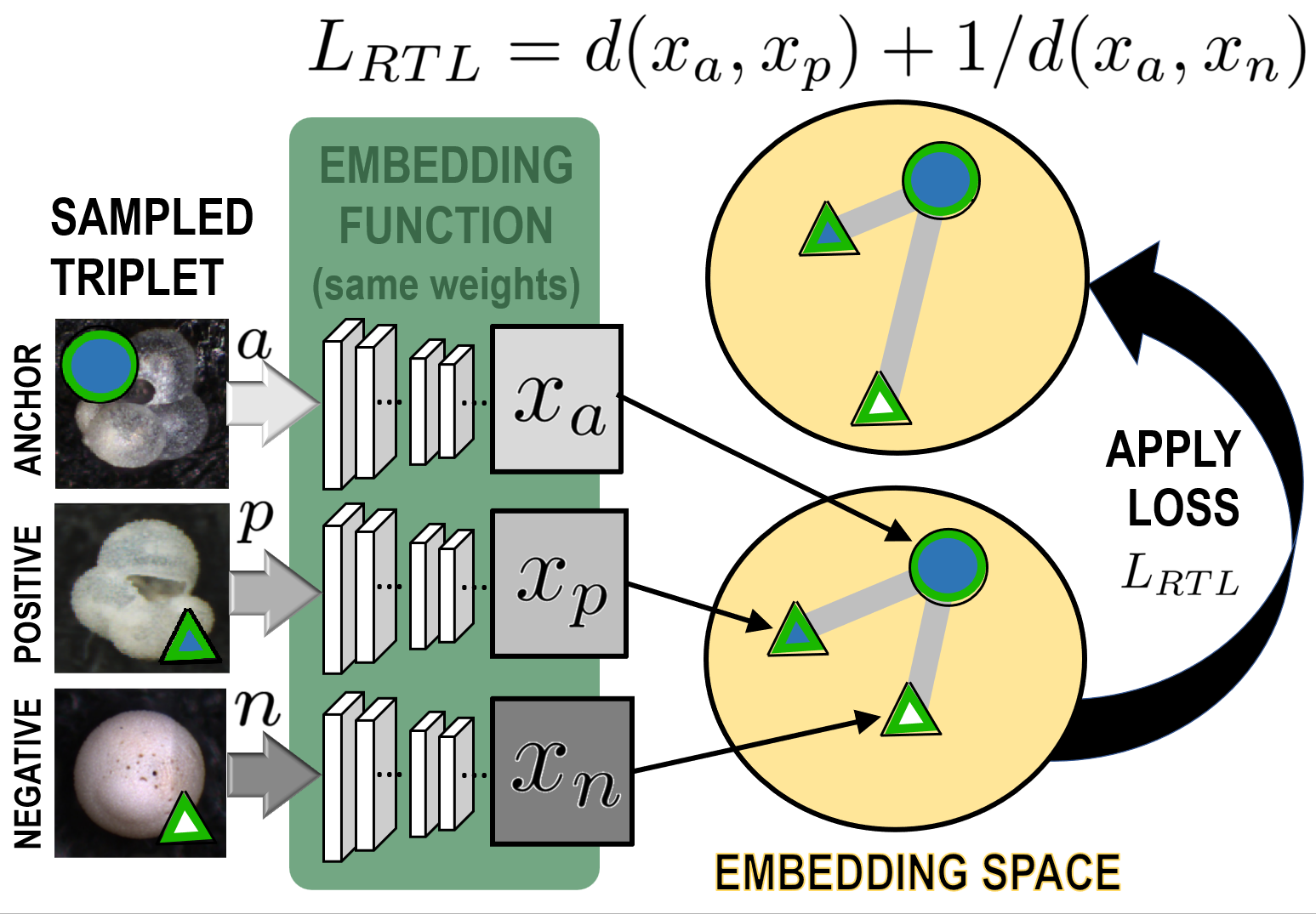}\vspace{-8pt}
  \end{center}
  \caption{ \textbf{Metric Learning via Reciprocal Triplet Loss.} Using triplets of an anchor sample (disc), a sample of the same (blue core triangle), and a sample of a different class (white core triangle) yields three vectors $x_a$, $x_p$, and $x_n$. The parameter-free loss function $L_{RTL}$ adjusts {Eucledian} distances $d(\cdot,\cdot)$ in embedding space by shortening $d(x_a,x_p)$ and lengthening $d(x_a,x_n)$ and can in conjunction with other losses~\cite{Tayfun} yield state-of-the-art results for learning species spaces directly usable for inference.}\vspace{-7pt}
  \label{triplet-figure}
\end{figure}

\subsection{Microfossil Classification via Metric Learning}\vspace{-3pt}
\textbf{Metric Learning.} The task of learning a classification-relevant similarity function from data is known as Metric Learning~\cite{andrew21,brookes23,gao21,schneider19, stennett22}, that is creating an embedding function that produces feature vectors in a space where samples of the same class cluster together far away from other data. Back-propagation with contrastive or triplet losses in the mix of cost functions~\cite{andrew21,Tayfun} can effectively implement such a system. The resulting distance metric can be used to perform tasks such as classification for both open or closed-set scenarios~\cite{andrew21}, clustering, and 
 retrieval~\cite{IR}. 

\textbf{Deep Microfossil Classification.} Recent taxonomic applications~\cite{Tayfun} of deep metric learning to visual microfossil identification achieve SOTA performance beyond other CNN approaches~\cite{endlessforams1,marchant} when evaluated on the large Endless Forams dataset~\cite{endlessforams1}. Reciprocal Triplet Loss, as illustrated in Fig.~\ref{triplet-figure}, together with the SoftMax Loss form the key cost functions used in these SOTA imageomics systems~\cite{Tayfun}. By combining the two losses both class-relative and class-absolute information can be utilised during learning. However, we note that the principal concept of adjusting distances via an anchor and nearby samples is not bound to a single modality such as vision. Instead, it provides an opportunity to transfer information across modalities by mixing anchor and sample modalities for alignment of different domains within one space~(see Section~\ref{alignment}).

\subsection{Long-Tailed Recognition}\vspace{-3pt}
 \textbf{Natural Data Collections.} The distribution of most biological datasets including microfossil~\cite{endlessforams1} datasets and other natural image collections~\cite{perrett2023,inaturalist} is long-tailed, that is a few classes have a lot more data than many other classes. This uneven distribution causes most machine learning models to perform poorly on the many rare classes. 
 
  \textbf{Specific Long Tail Techniques.} Long-tailed Recognition (LTR) techniques are used to improve the performance of models with a focus on rare classes. Different LTR methods have been proposed, such as re-sampling the training data to balance class distributions~\cite{resample1, resample2}, re-weighting classes and individual training examples~\cite{reweight}, transferring feature representations~\cite{face_rec} from common classes to rare classes, relating head and tail information~\cite{perrett2023}, decoupling feature learning and classifier learning~\cite{decouple1, decouple2}, or using self-supervised or ensemble models~\cite{ensemble}. For a more comprehensive overview of LTR, refer to the survey paper~\cite{LTR_review}. Since our target data are of taxonomic nature and contain rare classes LTR techniques offer a tool to potentially improve performance. In addition, this allows us to separate LTR compensation from the effect achieved by integrating genetic information. Thus, in this work, we explicitly utilise {weight balancing~\cite{LTRweightbalancing}, one of the latest state-of-the-art regularization approaches to LTR making use of weight decay and Max-Norm constraints} ~(see Section~\ref{LTR} for its impact).

\subsection{Genetic Data Embedding}\vspace{-3pt}
\textbf{Complexity of Genetic Data.} In order to integrate genetic information into other spaces sequence data needs to be placed or `embedded' within them.  Genetic sequence embedding is a challenging task due to the structuredness of potentially unaligned sequences of arbitrary length and content. In addition, a good embedding function for sequences has to capture both short- and long-term dependencies between symbols in the sequences.  

\begin{figure}
  \begin{center}
    \includegraphics[width=85mm,height=55mm]{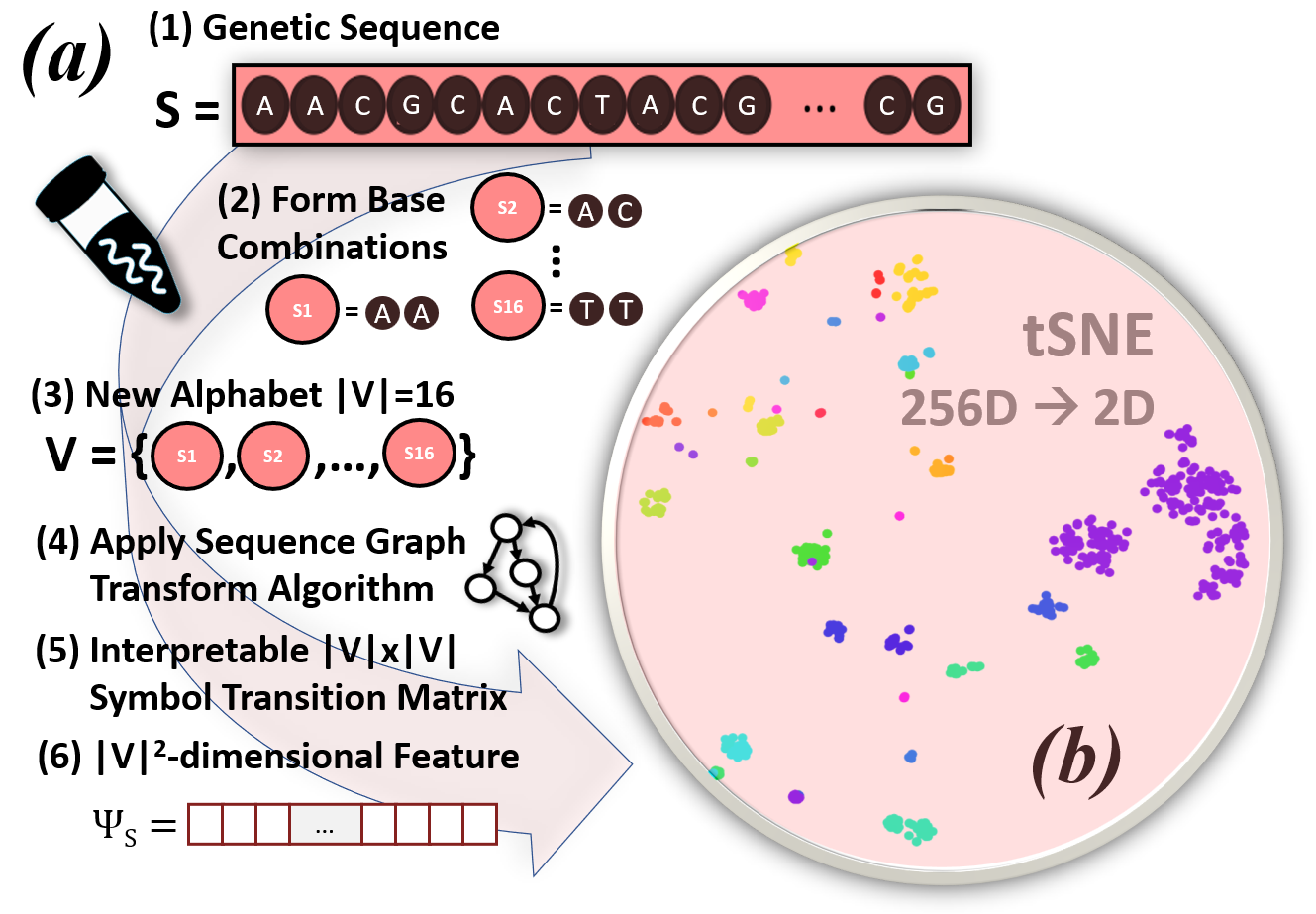}\vspace{-10pt}
  \end{center}
  \caption{\textbf{Genetic Embeddings via Sequence Graph Transform.} \textbf{\textit{(a)}} After creating an alphabet~$V$ of 16 symbols from pairs of the 4 bases, we use the Sequence Graph Transform to map a given genetic sequence, $S$ into a feature vector $\Psi_S$ of size $|V|$$\times$$|V|$. This vector is used as embedding and can also be interpreted as a graph that captures both long and short-term interactions between different symbols of the sequence. \textbf{\textit{(b)}}: 2D t-SNE visualisation of rDNA sequence embeddings for the 32 microfossil species.}\vspace{-8pt}
  \label{sgt-figure}
\end{figure}

\textbf{Embedding of Genetic Sequences.} For this task, Ranjan et al.~\cite{SGT} proposed the approach of a Sequence Graph Transform (SGT), a technique that represents sequences via the statistical relationships between symbols and casts this information into a feature vector. We opt to utilise this approach for creating embedding functions since it captures both local and global patterns into fixed-dimensional embedding vectors. In addition, the produced features are indeed interpretable where components represent directional dependencies between symbol pairs. 
Fig.~\ref{sgt-figure} visualises key features of the SGT technique. Note that, the grouping of pairs of symbols towards a new alphabet ensures the embedding vector is not dimensionality-limited -- despite the fact that the square of symbol cardinality is the fixed embedding vector dimension. Section~\ref{alignment} details the procedure followed for genetic data in this work.

\subsection{Cross-Domain Transfer Learning}\vspace{-3pt}
\textbf{Knowledge Carry-over.} Applying knowledge gained from one domain to a different -- but related -- domain for improved performance is known as Cross-Domain Transfer Learning. Such a methodology is clearly most useful if data is limited in the target domain itself. Since long-tailed taxonomic image collections exhibit exactly this property, we propose to transfer knowledge learned from the genetic domain into the visual domain.

\textbf{Vision-Language Transfer.} Classic transfer learning approaches in the literature often focus on bridging image and text domains. They include DeViSE (Deep Visual-Semantic Embedding)~\cite{device} to perform zero-shot image classification via Word2Vec~\cite{word2vec} embeddings used to link visual and language-based semantics. Karpathy et al. proposed VSA (Deep Visual-Semantic Alignments)~\cite{VSA} for this task, which uses R–CNN~\cite{RCNN} and BRNN~\cite{BRNN} to extract image features and text features, respectively. Kiros et al. proposed UVSE (Unifying Visual-Semantic Embedding)~\cite{UVSE}, which uses a VGG-19~\cite{VGG} network to extract image features and an LSTM~\cite{LSTM} network for text. This was later extended to include hard negative mining in VSE++~\cite{vse++}. 

\textbf{Concept of Compatible Representation.} One key feature of most approaches is the mapping of different domains into a common data format where information transfer becomes possible. Our cross-modality learning follows this concept by mapping visuals via ResNet50 and rDNA information via SGT into a common space. We  use settings similar to DeViSE for cross-domain training.

%\textbf{Alignment Mechanism. \tayfun{DeViCE introduces an alignment mechanism that establishes a shared embedding space for visual and semantic features. It utilises CNNs to obtain visual descriptors, and Word2Vec for semantic descriptors. In the alignment phase, the late projection layers of the CNN model are adjusted so that the visual desriptors become more aligned with their corresponding semantic descriptors in the shared embedding space. This allows the model to capture the relationships between images and their textual descriptions. In this work, with inspiration from DeViSE, we utilize CNNs for image representations and SGT for genetic representations. Similarly to DeViCE, in the alignment phase, we adjust the late visual representations so that they become more aligned with their corresponding genetic descriptors as in Fig.~\ref{title-figure}. This allows the model to capture, cross-domain associations between genetic and morphology spaces, for a more generalized understanding of the species concepts with a prior understanding of what the visual of a species may look like based on their genetic similarities to other species.}}

\textbf{Alignment Mechanism.} {DeViCE establishes a shared embedding space for visual and semantic features, originally utilizing CNNs for visual descriptors and Word2Vec for semantic descriptors. For alignment, the CNN's late projection layers are trained to align visual descriptors with semantic ones in a shared space, originally capturing image-text relationships. Inspired by DeViCE, we use CNNs for image representations and SGT for genetic ones. Similarly, we train projection layers of the CNN to align visual descriptors with genetic ones, as depicted in Fig.~\ref{title-figure} and Fig.~\ref{tl-figure}.}

\begin{figure*}[t]
  \begin{center}\vspace{-5pt}
    \includegraphics[width=500pt,height=240pt]{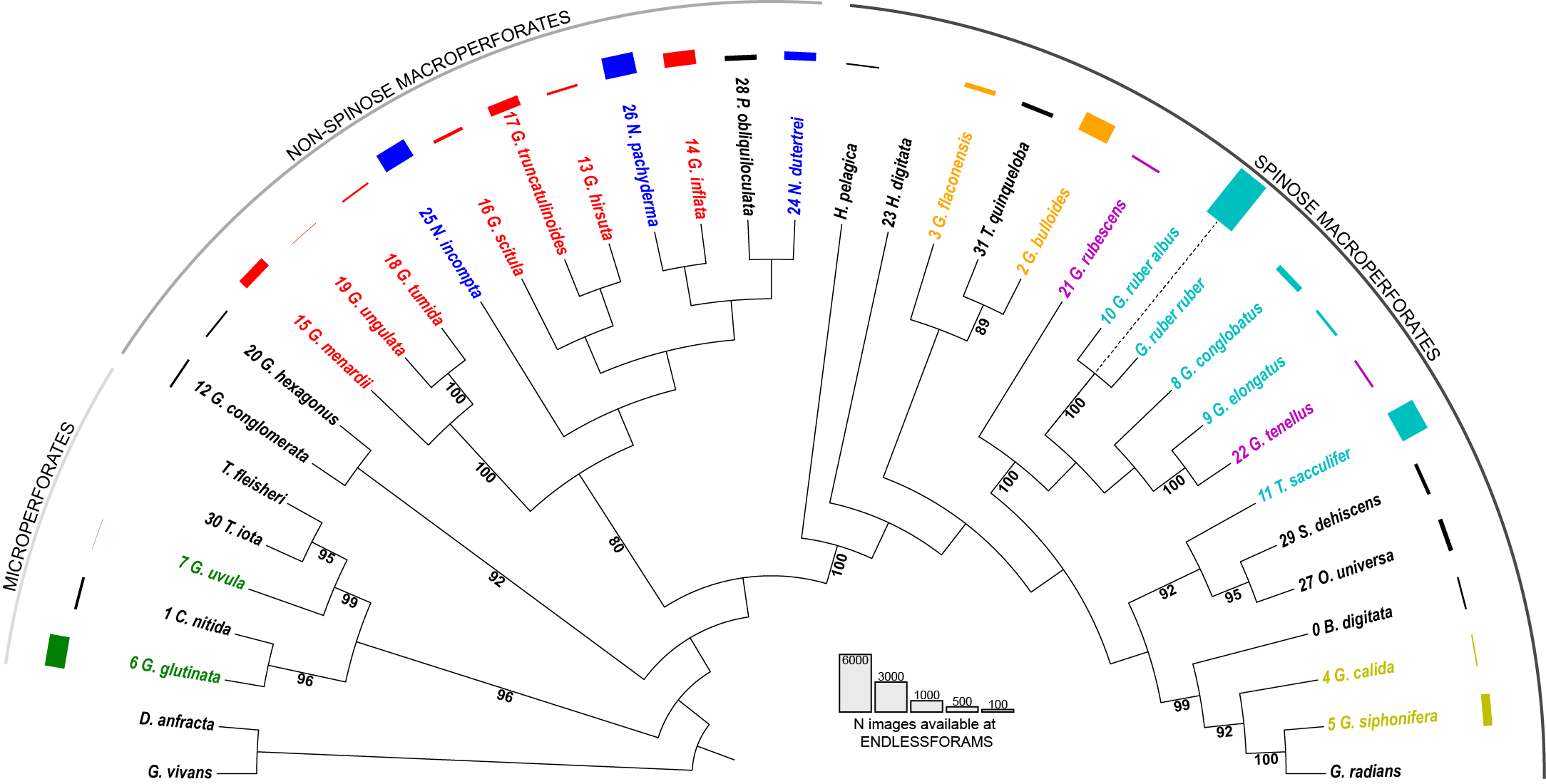}\vspace{-10pt}
  \end{center}
  \caption{\textbf{Phylogenetic Reconstruction of Relevant Foraminifera.} RAxML phylogenetic inference showing the relationships between the extant planktonic foraminifera without branch length. The values next to the branches indicate the bootstrap values and the bars next to the species names the number of images used in the study. Numbers 0 to 31 associated with the taxa used in experiments align with the visual depictions in Fig.~\ref{spaces-figure} and the detailed histogram in Fig.~\ref{taxa-figure}. The tree is rooted on the phylogenetically basal \textit{G. vivans} and \textit{D. anfracta}.}\vspace{-5pt}
  \label{tree-figure}
\end{figure*}

\vspace{-6pt}
\section{Datasets}\vspace{-4pt}
\subsection{Endless Forams Imagery}\vspace{-4pt}
\textbf{Taxa with Visual-Genetic Support.} We utilise 32 out of 35 species from the public Endless Forams image library~\cite{endlessforams1} for all our experiments. That is we utilise exactly those species for which we were able to gather sufficient genetic rDNA information. Figure~\ref{tree-figure} shows an associated phylogenetic tree. Endless Forams is one of the largest datasets of its kind and freely available at~\url{endlessforams.org} supporting full reproducibility of our experiments. In its entirety, it contains 34,640 species-labelled and location-centred images of 35 different marine calcareous plankton species (foraminifera) as detailed phylogenetically in Fig.~\ref{tree-figure}. Further, Fig.~\ref{spaces-figure} depicts the covered taxa and Fig.~\ref{taxa-figure} shows a histogram of sample sizes. 

%\textbf{Challenging Imagery.} Species classification in this data is challenging for machines since  foraminifers exhibit significant intra-class variability and the dataset is highly unbalanced with sample cardinatities ranging from $7$ to $5,914$ per taxa. Moreover, even manual taxonomic classification is complicated since critical morphological properties are not consistently imaged across 3D viewpoints, acquisition conditions vary, and out-of-focus problems may hide details.

\subsection{Endless Forams Genetics}\vspace{-3pt}
\label{genetic}
\textbf{Genetic Data Sources.} For genetic information, we use a fragment of the 18S rDNA sequences (ribosomal genes DNA) of the 32 species selected in the study to serve as embeddings. We select the sequences from the Planktonic Foraminifera Reference Database~\cite{morard2015pfr2} publically available at~\url{http://pfr2.sb-roscoff.fr, } and included sequences published afterwards~\cite{morard2019genetic,morard2019unassigned,ujiie2016evolution}. Overall, we retained 878 sequences covering the entire barcode selected for foraminifera studies~\cite{pawlowski2014plea} and provide the sequence list as supplementary material. 

\textbf{Pylogenetic Tree Inference.} As discussed,  Figure~\ref{tree-figure} reconstructs the evolutionary relationships between extant species via phylogenetic inference from sequence representation to a tree structure. We included further species to cover the full phylogenetic spectrum of planktonic foraminifera in this representation. The sequences were automatically aligned using MAFFT v.7~\cite{katoh2013mafft}, and we inferred the phylogenetic tree with RAxML-NG~\cite{kozlov2019raxml} using the model GTR+I+G that was selected with Modeltest-NG~\cite{darriba2020modeltest} and with 100 non-parametric bootstrap runs. 

\section{Generating Visual-Genetic Spaces}\vspace{-3pt}
\subsection{Metric Visual-Only Pre-Training.}\vspace{-3pt}
\label{pretrain}
\textbf{Constructing Latent Image Embeddings.} In order to construct a maximally rich initialisation of a task-aligned data space before visual-genetic integration we first construct a traditional deep mapping from source images to a class-distinctive embedding space~\cite{Tayfun}. The simplest way of creating such a metric embedding is via the use of a basic contrastive loss~$L_{C}$~\cite{20} using pairs of data points:
\begin{equation}
L_{C} = \frac{(1-Y)}{2}d(x_{1}, x_{2})+\frac{Y}{2}max(0, \alpha - d(x_{1}, x_{2})),
\end{equation}

\noindent
where $x_{1}$ and $x_{2}$ are the embedded input vectors, $Y$ is a binary label denoting class equivalence/difference for the two inputs, and $d(\cdot,\cdot)$ is the Euclidean distance between two embeddings. However, this formulation cannot put similarities and dissimilarities between different embedding pairs in relation. A triplet loss formulation~\cite{21} instead utilises three embeddings $x_a$, $x_p$ and $x_n$ denoting an anchor, a positive example of the same class, and a negative example of a different class, respectively:
\begin{equation}
L_{TL} = max(0; d(x_a,x_p) - d(x_a,x_n) + \alpha), 
\end{equation}
where $\alpha$ is the margin hyper-parameter. Reciprocal Triplet Loss as visualised in Figure~\ref{triplet-figure} removes the need for this parameter~\cite{22} and naturally accounts for offsetting the impact of large margins far away from the anchor:
\begin{equation}
L_{RTL} = d(x_a,x_p) +1/d(x_a,x_n).
\end{equation}
As shown by recent work~\cite{23,24}, including a SoftMax term in this loss can improve performance further. Thus, SoftMax and Reciprocal Triplet Loss can be combined into a standard formulation  used here and published in~\cite{andrew21} as a mixture with balancing hyper-parameter $\lambda$:
\begin{equation}
\label{loss}
L = -log(\frac{e^{x_{class}}} {\sum_i e^{x_{i}}}) + \lambda L_{RTL}.
\end{equation} 

\textbf{Application-Specific Relevance and Baseline.} For the foraminifer classification problem at hand this allows for the use of both relative inter-species information captured by the $L_{RTL}$ component as well as absolute species information captured by the SoftMax term as back-propagation gradient components. Training a latent embedding space as described essentially acts as a single modality baseline~(see Fig.~\ref{title-figure}), replicating a SOTA image-only deep learning solution following~\cite{Tayfun} to solve the biometric species identification problem. 

\subsection{Genetic to Visual Information Transfer}\vspace{-3pt}
\label{alignment}
\textbf{Multi-Symbol Genetic Embedding.} rDNA sequences as resulting from genetic sources as described in Section~\ref{genetic} strictly contain four base symbols, that is ${A, C, G, T}$. Thus, the SGT algorithm applied to this raw data would produce an embedding vector of very low dimensionality, that is $4 \times 4 = 16$. For rDNA vectors to structurally fit the high-dimensionality required for visual embeddings, we therefore bin every two symbol terms in rDNA sequences to form a new alphabet of $4 \times 4 = 16$ symbols made of $4$ base symbols. Application of SGT to this new symbol set then creates embeddings of size $16 \times 16 =256$ as required. Figure~\ref{sgt-figure} depicts a 2D t-SNE representation of the resulting genetic embedding space. Note that the embedding dimension may be adjusted to some lower size via compression via a trained fully connected or convolutional layer. 

\begin{figure}[t]
  \begin{center}
    \includegraphics[width=85mm,height=55mm]{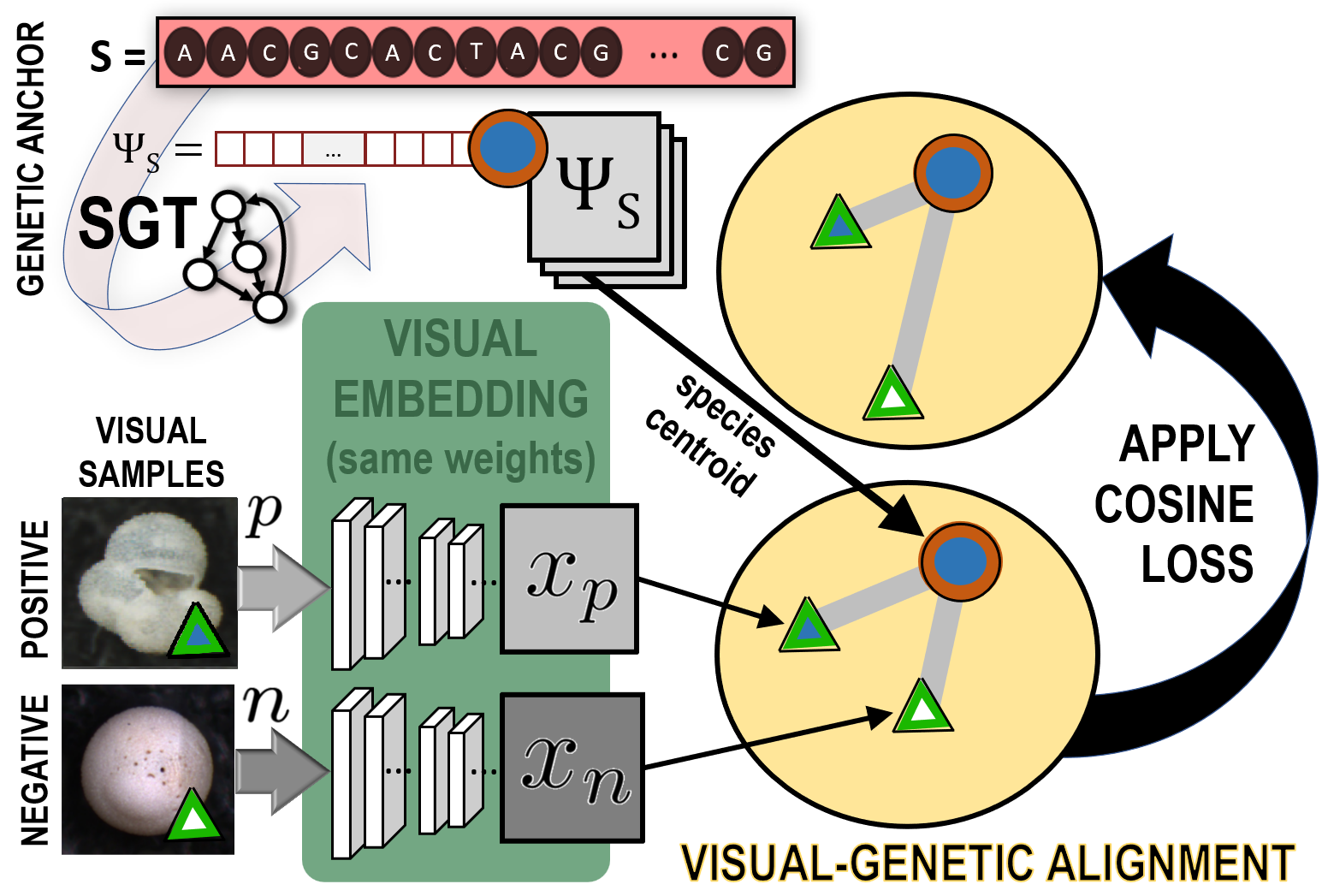}\vspace{-5pt}
  \end{center}
  \caption{\textbf{Visual-Genetic Alignment via Transfer Learning.} The proposed cross-modality alignment of pre-trained ResNet50 visual embeddings towards genetic anchors generated via SGT uses a mixed-modality triplet cosine loss to transfer information. See Fig.~\ref{triplet-figure} for further details regarding symbol semantics.}\vspace{-10pt}
  \label{tl-figure}
\end{figure}

\textbf{Visual Model Alignment Towards Genetics.} The visually pre-trained ResNet50 model (see Section~\ref{pretrain}) is used as initial embedding function. Since the dimensional structure of this space is identical to the rDNA embeddings, the latter can be used to guide visual embedding positions. For each species a single rDNA target embedding is calculated as a genetic anchor defined as the median vector over all available rDNA embeddings of the taxa. {Note that the choice of median vector over the mean vector is to eliminate the impact of any outliers that may be present in the datasets.} Given this, we freeze the convolutional layers of the ResNet50 model and tune the remaining projection layers to capture cross-domain information. Methodologically, this is achieved by using a triplet formulation that uses Cosine distances. That is, for anchor-positive pairs the loss is defined as $1 - cos(x_1,x_2)$, while for anchor-negative pairs, the loss is defined as $max(0, cos(x_1,x_2))-m)$ and these two terms are summed to form a loss $L_{Cosine}$. The parameter $m=0.5$ is the margin set as recommended in \cite{NEURIPS2019_9015}. Repeated application of this loss moves the model towards a higher orientational alignment between visual model and genetic anchors transferring information from the latter towards the visual imageomics classification model.

%\textbf{Choise of Distance Metric}. \tayfun{To give context, the cosine distance is often preferred over the Euclidean distance for aligning vision and language data for several reasons. First, cosine similarity's normalization of vectors makes it robust to varying scales in different modalities. Second, by emphasizing directional alignment rather than magnitudes, it captures semantic relationships crucial in cross-modal tasks. Third, it's slightly less affected by the curse of dimensionality, making it suitable for high-dimensional data. Fourth, the loss aligns with the goal of cross-modal understanding, promoting semantic similarity. Lastly, its translation invariance and focus on semantic relationships make it well-suited for aligning features in vision and language alignment tasks. Similarly, in the context of aligning vision and genetic data, our experimental findings highlight the advantages of employing cosine distances within our triplet formulation to derive the cosine embedding loss. This approach demonstrated notable enhancements in tail-class performance. In contrast, employing distances associated with the standard $L_{RTL}$ loss for alignment purposes resulted in performance declines. These outcomes reaffirm our anticipations, underlining the similarity between the machine learning task of aligning vision and genetic data and the established parallels with vision and language alignment.}
\textbf{Choise of Distance Metric}. {Cosine distances operate invariant to vector scaling, ie. their application is implicitly scale-normalized which is critical for information transfer between non-aligned spaces. Experimentally, our triplet formulation of the cosine embedding loss boosted tail-class performance when compared to using standard Euclidean distances $L_{RTL}$ for alignment.}

\begin{figure*}\vspace{-9pt}
  \begin{center}
    \includegraphics[width=172mm,height=75mm]{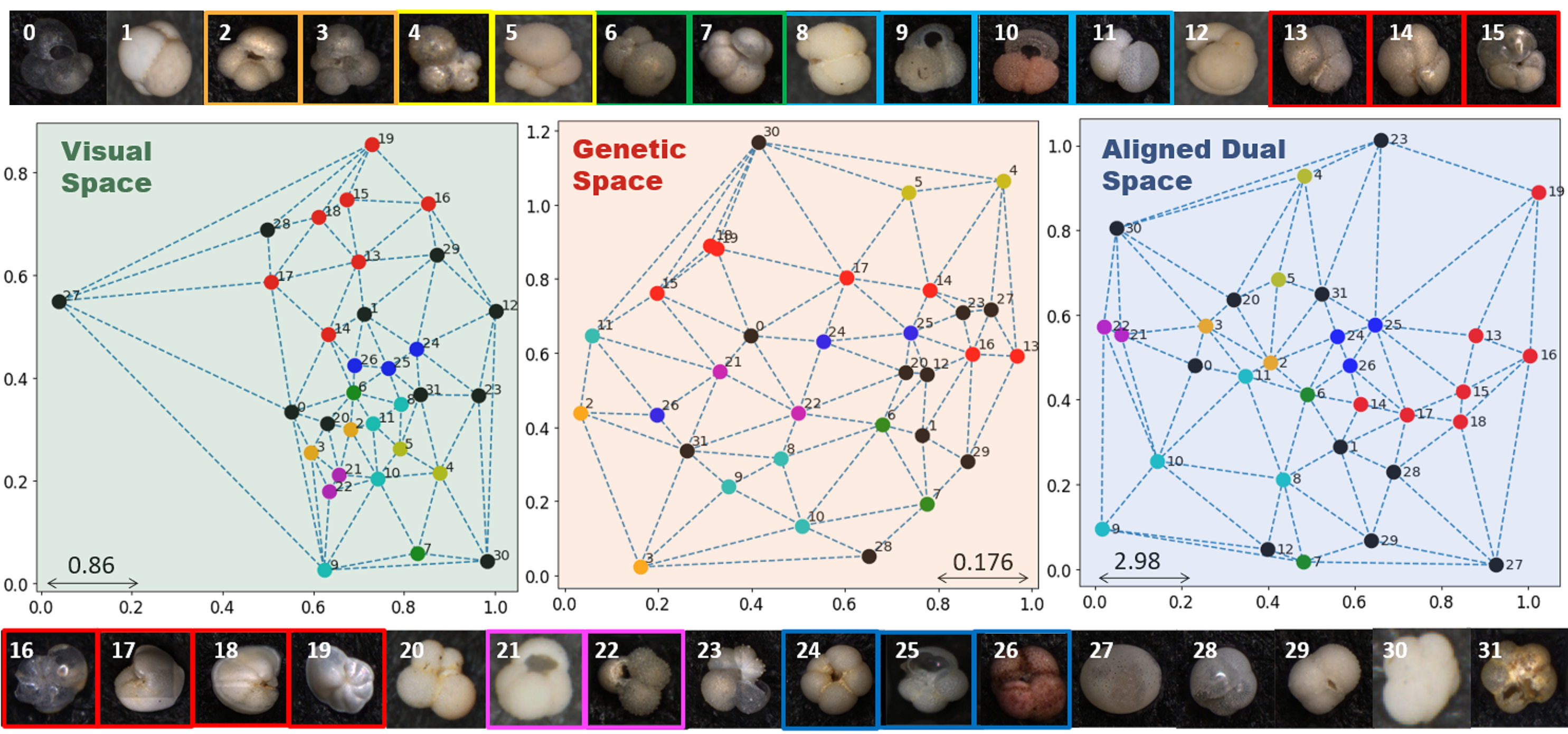}\vspace{-10pt}
  \end{center}
  \caption{\textbf{Structure of Latent Species Spaces.} Visual samples of all 32 numbered taxa together with a 2D visualisation of Delaunay Triangulations of the taxa centroids for three latent spaces constructed: \textbf{\textit{(left)}} visual-only baseline space, \textbf{\textit{(middle)}} SGT-created genetic space, and \textbf{\textit{(right)}} our proposed visual-genetic dual space still permits for visual-only inference and reveals qualitatively improved grouping of genera (shown by colours) together with more equidistant taxa spacing compared to the visual model resulting in superior per-class inference. Note that 256D latent spaces are approximately visualised in 2D by taking cosine distance matrices and minimizing a global energy function via the Kamada Kawai~\cite{kamada1989algorithm} algorithm. Different colors (excluding black) correspond to different genera in line with Fig.~\ref{taxa-figure}.}\vspace{-5pt}
  \label{spaces-figure}
\end{figure*}

\begin{table*}[t]
  \begin{center}
    \includegraphics[width=500pt]{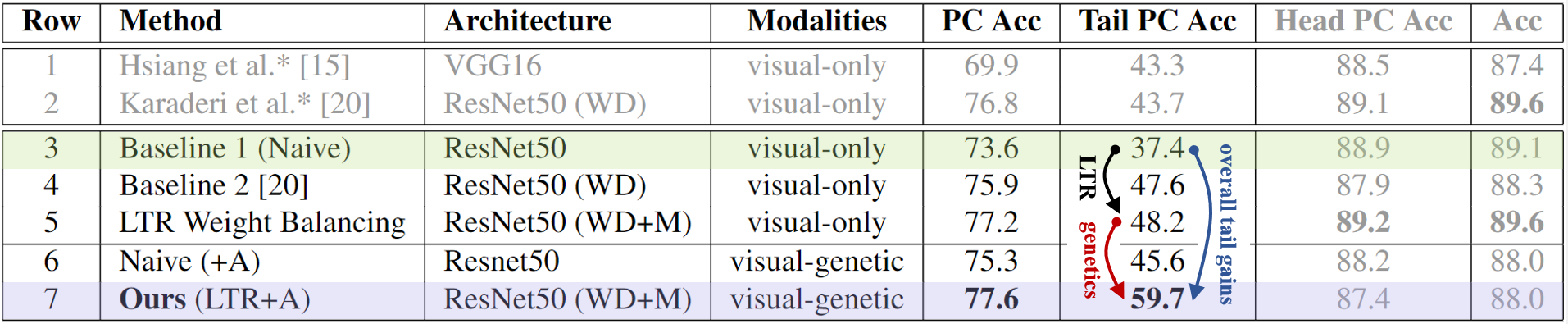}\vspace{-10pt}
  \end{center}
\caption{\textbf{Quantitative Results with Focus on Per-Class and Tail Performance.} Experimental accuracy (Acc) results for the Endless Forams visual-only test set across the 32 used taxa at 8bit grayscale and $160 \times 160$ pixels resolution. Starting from a Naive visual vanilla model (row 3) adding LTR weight balancing here for the first time to the domain (row 5) improves all benchmarks and beats the SOTA Baseline 2 model (row 4). Further adding our proposed visual-genetic alignment to this LTR training (row 7) boosts tail performance to $59.7\%$ and can further enhance per-class accuracy at a cost of only $1.6\%$ overall accuracy loss. [Tail Per-Class (PC) accuracy is for classes with less than 100 visual samples. Head PC accuracy is for classes with more than 1,000 samples.] [A: visual-genetic alignment, WD:~Weight Decay, M: Maxnorm. *: not directly comparable models with access to 35 visual training 
 classes.]}
 \label{tabres}
\end{table*}

\section{Experimental Setup}\vspace{-3pt}
\subsection{Implementation Details}\vspace{-3pt}
\textbf{Basic Training Details.} For all experiments, we use a PyTorch-implemented metric learning architecture with a ResNet50 backbone pre-trained on ImageNet~\cite{deng2009imagenet} using two fully connected projection layers to map the standard ResNet50 feuture space of width 2048 to first 1000 and then to our 256-element feature vector, which is the same embedding size as for the genetics. For universal comparability, we utilise a fixed, withheld test set of $6,801$ images for performance stipulation, whilst using the remaining $27,731$ images augmented via rotations, scale, and Gaussian noise transforms for training. Exact sample-accurate data splits are published with this paper for full reproducibility. The network is first tasked to optimize the visual-only loss specified in Eq.~\ref{loss} combining SoftMax and Reciprocal Triplet Loss components with the mixing parameter $\lambda=0.01$~\cite{Tayfun} as described in Section~\ref{pretrain} via SGD for 20~epochs. We train both a naive baseline version and one enhanced with LTR weight balancing~\cite{LTRweightbalancing} to separate the effect of LTR learning from genetic alignment. {Our published source code~\cite{repo} provides full details regarding all of the above and result reproducibility.}

\textbf{Cross-Modality Alignment.} After these 20 epochs of visual-only training, we engage our genetic anchors obtained through SGT and align the space across modalities {using a Cosine embedding loss with triplet formulation} as explained in Sec.~\ref{alignment} for another 5 epochs via SGD. We note that the choice of this loss function is critically important when integrating visual-genetic spaces. Neither using SGT embeddings as anchors with the standard Euclidean-based $L_{RTL}$ loss nor utilizing extra trainable integration layers yielded any integrative success and cross-modality data transfer. Consequently, the use of the proposed orientational cosine alignment between visual embeddings and genetic anchors~(see Fig.~\ref{tl-figure}) proved a critical design choice and effective cost function for information transfer. Full implementation details are available via the source code repository published with this paper. 

\section{Results and Discussion}\vspace{-3pt}

\subsection{Quantitative Performance Benchmarks}\vspace{-3pt}
\label{LTR}

\textbf{Baselines.} Table~\ref{tabres} quantifies performance for the system and across various baselines. We show performance of the original Endless Forams paper in \textit{row~1} and the current SOTA work in \textit{row~2} recalculating performance for our 32 covered taxa from the exact models built from all 35 taxa. For fair comparability we then retrain the SOTA model in \textit{row 4} on only our 32 taxa covered (ie. where genetic data was available). For full ablation of effects we also stipulate performance for the exact model used, but without genetic alignment and without LTR techniques given as 'Naive' model in \textit{row~3}. All baselines show low performance on rare tail classes compared to per-class or general accuracy.

%\begin{figure}[b]
%  \begin{center}\vspace{-5pt}
%    \includegraphics[width=85mm,height=23mm]{figures/visualt.png}\vspace{-10pt}
%  \end{center}
%  \caption{\textbf{Losses and Accuracy during Visual Pre-Training.} Losses~(red) and validation accuracy~(green) during 20 epochs of visual pre-training with LTR weight balancing. Learning saturates with limited training potential left for visual-only optimisation.}
%  \label{train-figure}
%\end{figure}

\begin{figure}[t]
  \begin{center}\vspace{-10pt}
    \includegraphics[width=82mm,height=72mm]{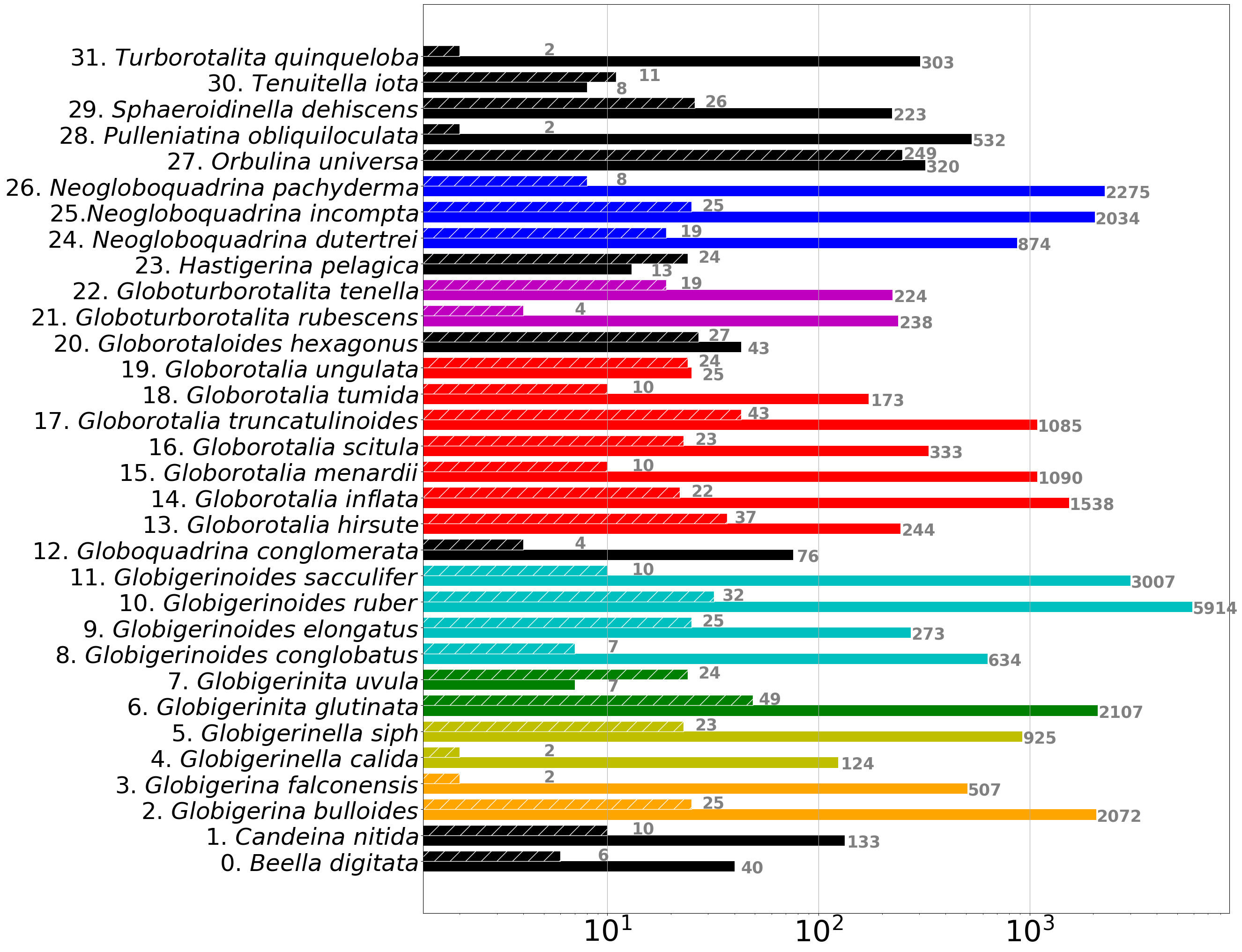}\vspace{-10pt}
  \end{center}
  \caption{\textbf{Histogram of Available Samples Across Taxa.} A visualisation of the highly variable sample number associated to the 32 taxa used in the study (Dashed Upper Bars: rDNA, Filled Lower Bars: Visual). Groupings of taxa known as genera are indicated by colouration in line with Fig.~\ref{spaces-figure}.}
  \label{taxa-figure}
\end{figure}

\textbf{Improving SOTA Performance via LTR.} This work is the first to apply a SOTA LTR technique to the problem at hand. When adding weight balancing~\cite{LTRweightbalancing} in~(row~5) to the 'Naive' Baseline 1, we were able to outperform all benchmarks and also beat the directly comparable SOTA Baseline 2~(row~4) across all measures without degradation of overall top accuracy. However, without using genetic information tail performance gains of LTR techniques are still limited. 

\textbf{Visual-Genetic Alignment.} When adding visual-genetic alignment techniques to the 'Naive' Baseline 1 without LTR techniques (see \textit{row~6}) an overall increase in per-class accuracy of $1.7\%$ can be seen and for rare classes (N$<$100) per-class accuracy improved significantly by $8.2\%$. However, in order to improve performance for rare classes further we tested combining visual-genetic alignment with LTR-improved latent spaces.

\textbf{Improving Per-Class and Rare Class Benchmarks.} Combining the proposed visual-genetic alignment with LTR training (row~7) boosts tail performance significantly by another $11.5\%$ to $59.7\%$ and can further enhance per-class accuracy at only $1.6\%$ overall accuracy trade-off. Given the practical importance of determining the presence of rare classes in paleoclimatology\cite{diversity} this cost at the distribution's head is negligible in many important tasks. Moreover, the effect of genetically-driven improvements is particularly pronounced for the rarest classes where least image data is available. Fig.~\ref{samplesize-figure} plots per-class accuracy against cardinality of available class samples. This analysis confirms that a visual-genetic LTR model consistently yields superior performance to component ablations and that relative gains grow with sample rarity.

\textbf{Visual-Genetic Alignment Control.} In order to show that genetic information indeed has a domain-specific effect, we also test visual-genetic alignment with a ResNet50 model trained only on the ImageNet, that is one which has never seen images of foraminifers. We aligned this model as before to genetic anchors. As shown in Fig.~\ref{dist-figure}\textit{(a)}, such alignment still improves both overall accuracy (by $16.9\%$) and the per-class accuracy (by $15.5\%$) from the generic off-the-shelf ImageNet pre-training model. This confirms that visual-genetic alignment indeed transfers domain-specific information.

\subsection{Qualitative and Class-Specific Discussion}\vspace{-4pt}
\textbf{Structure of Species Spaces.} Based on these results, Fig.~\ref{spaces-figure} explains and depicts a structural visualisation of the underlying latent spaces. It can be seen that superior per-class inference observed in the proposed visual-genetic dual space on the \textit{(right)} is underpinned by more equidistant taxa spacing compared to the visual-only Baseline 1 model on the \textit{(left)}. In addition, an improved clustering of genera (see colours) shows that aligned information from phenotype and genotype can indeed efficiently encode relationships between taxa in the metric structure of a latent space.

\begin{figure}
  \begin{center}
\includegraphics[width=80mm,height=45mm]{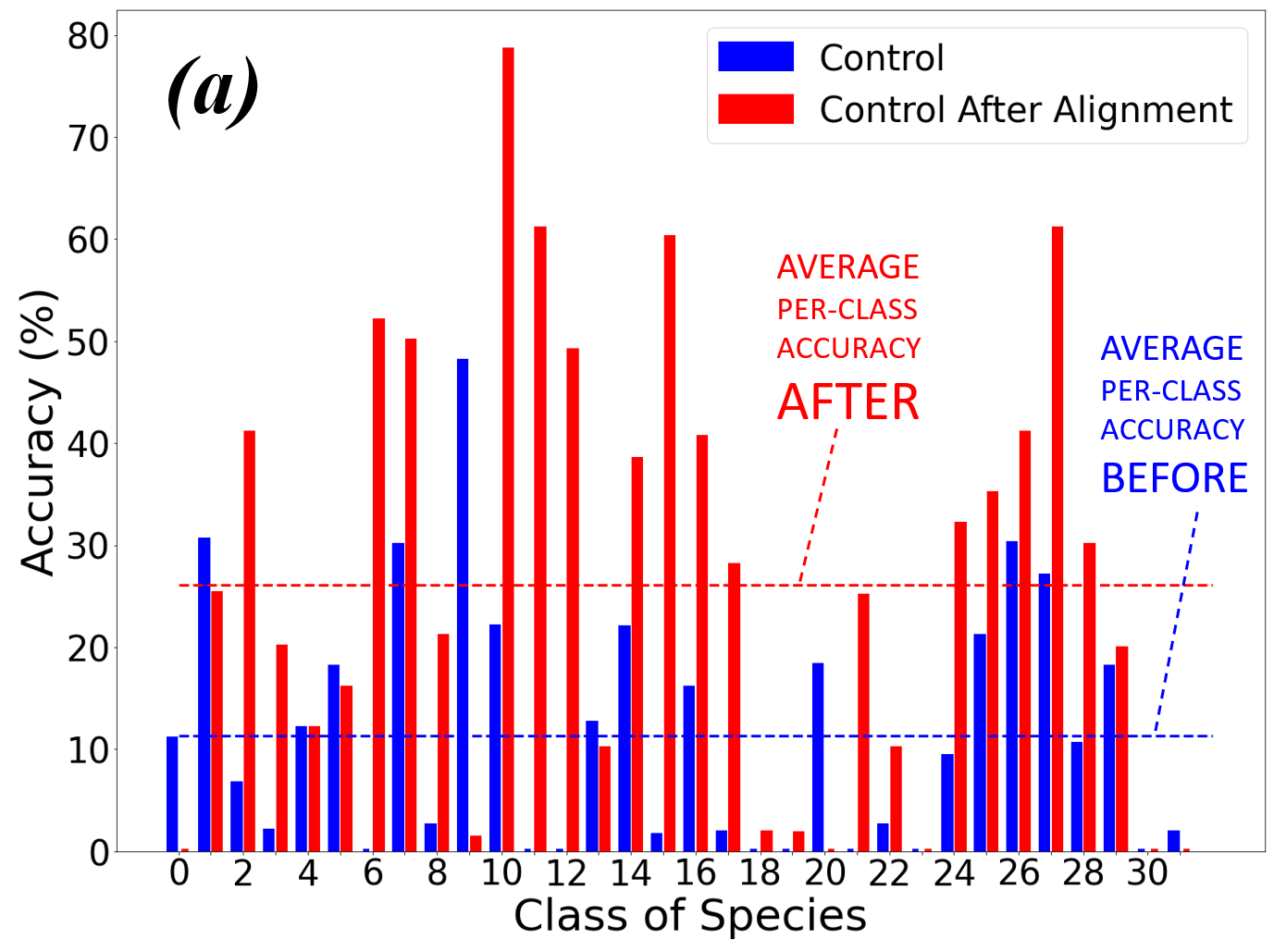}\vspace{9pt}
\includegraphics[width=80mm,height=45mm]{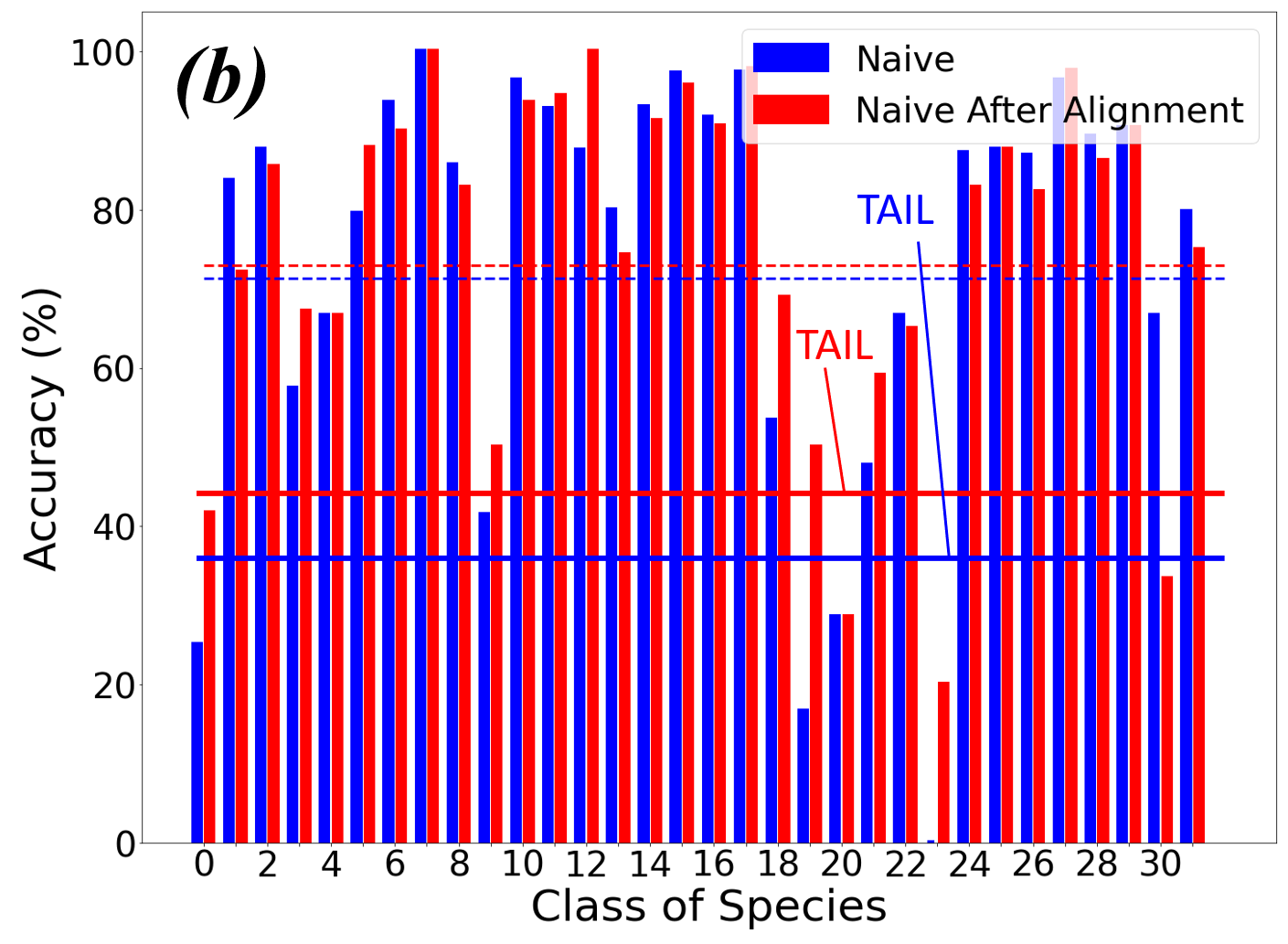}\vspace{9pt}
\includegraphics[width=80mm,height=45mm]{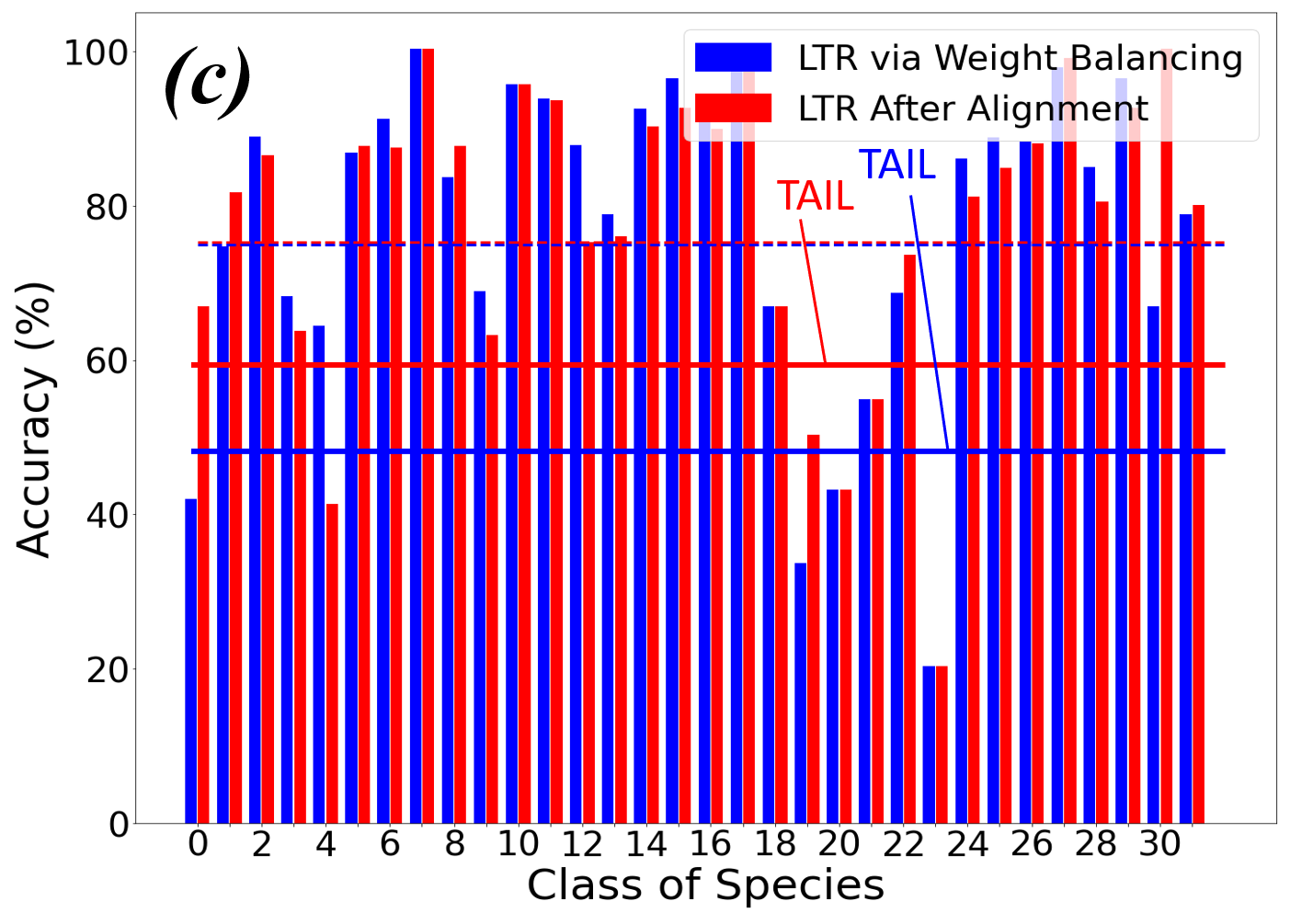}\vspace{-5pt}
  \end{center}
  \caption{\textbf{Alignment Effect on Per-Class Accuracy.} Visual-ge-netic alignment consistently improves visual-only per-class test accuracy particularly for tail classes. \textbf{\textit{(a)}}~Alignment improvements for a out-of-domain control network initialised on ImageNet highlight effective information transfer from the genetic domain; \textbf{\textit{(b)}}~Alignment of the Baseline 1 'Naive' visual-only model shows significantly raised tail performance; \textbf{\textit{(c)}}~Alignment of an LTR visual-only model shows further per-class accuracy improvements with high gains on tail classes well beyond LTR-only performance.\vspace{-9pt}}
  \label{dist-figure}
\end{figure}

\begin{figure}
  \begin{center}
   \includegraphics[width=84mm,height=25mm]{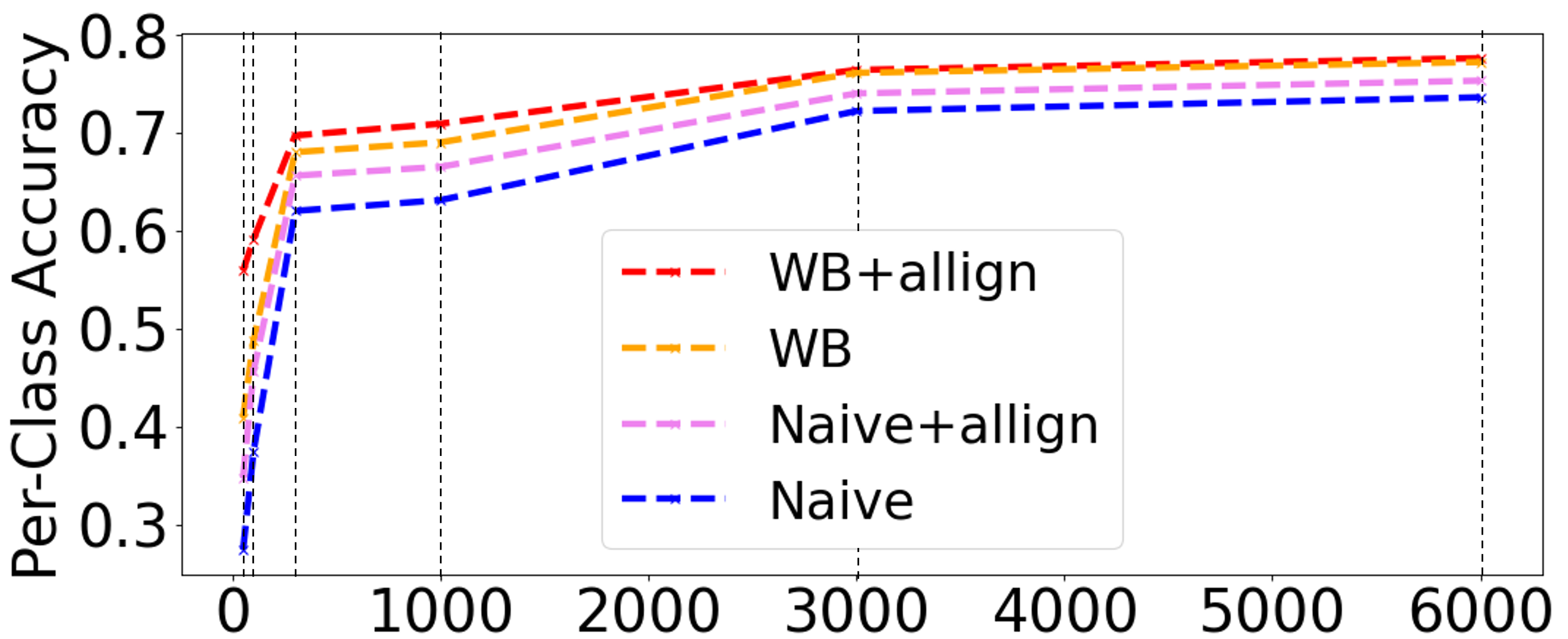}\vspace{-6pt}
  \end{center}
  \caption{\textbf{Tail-Class Performance vs Sample Abundance.} Average per-class accuracy for classes with samples less than the abscissa value. The proposed combination of LTR weight balancing and visual-genetic alignment outperforms other ablations noting that gains improve with sample rarity.}
  \label{samplesize-figure}\vspace{-15pt}
\end{figure}

\textbf{Taxa-Specific Discussion.} In order to quantify performance fine-grained on taxa level, Fig.~\ref{dist-figure} depicts a breakdown of the effect of transferring genetic information towards the visual domain for control, 'Naive' visual, and LTR enhanced models. For the rare tail classes {0,12,19,20,22,29,31} in particular accuracy increases after alignment for classes {0,19,22,29,31}. For the remaining two rare classes 12 and 20 the genetic placement on the phytologentic tree of foraminifers~(see Fig.~\ref{tree-figure}) is in fact unclear~\cite{mikrotax} according to domain experts, thus casting doubt over whether genetic information can at all be reliable for improving visual classification in these taxa. Overall, per-class accuracy \textit{consistently} improves after visual-genetic alignment with highest impact on tail classes quantified in Fig.~\ref{samplesize-figure}. Thus, the novel concept of proposed deep visual-genetic biometrics is demonstrably effective in the tested domain of taxonomic species classification.

\vspace{-4pt}
\section{Conclusion and Future Work}\vspace{-6pt}
\textbf{Visual-Genetic Biometrics for Rare Taxa.} We presented visual-genetic biometrics, a novel domain for improving visual taxonomic classification in datasets with rare species via information transfer from the genetic domain. We provided a deep proof-of-concept implementation that leverages rDNA data to align imageomic and genetic information to create a multi-domain embedding space. Using 30k+ visuals across 32 taxa from the Endless Forams dataset we first demonstrated that traditional CNN application can be enhanced by LTR techniques to outperform the state-of-the-art on all benchmarks. We then showed that visual-genetic alignment can further improve per-class performance, particularly for rare classes. This establishes a new benchmark and confirms the effectiveness of visual-genetic biometrics by proof-of-concept. We note that the latent species space built with LTR techniques is particularly receptive to genetic information transfer.

\textbf{Future Integration of Imageomics and Genetics.}
We believe that the implementation of artificial intelligence systems that organise life based on both phenotype and genotype will be important and significant with respect to reconciling genetic and imageomic spaces. This stems from a widely unexplored potential for a data-driven formal integration of the various approaches to the classification of life and for establishing tractable interfaces between the forms and levels life exhibits.

\vspace{-4pt}
\section*{Acknowledgements}\vspace{-6pt}\small
\noindent TK was supported by the UKRI CDT in Interactive Artificial Intelligence under
the grant EP/S022937/1. DNS was supported by NERC grant NE/P019439/1. Thanks also to Allison Y. Hsiang and the Endless Forams team.

{\small
\bibliographystyle{ieee}
\bibliography{egbib}
}

\end{document}